\documentclass[sigconf]{acmart}

\AtBeginDocument{%
  \providecommand\BibTeX{{%
    \normalfont B\kern-0.5em{\scshape i\kern-0.25em b}\kern-0.8em\TeX}}}

\setcopyright{acmcopyright}

\copyrightyear{2021}
\acmYear{2021}
\setcopyright{acmcopyright}\acmConference[MM '21]{Proceedings of the 29th ACM International Conference on Multimedia}{October 20--24, 2021}{Virtual Event, China}
\acmBooktitle{Proceedings of the 29th ACM International Conference on Multimedia (MM '21), October 20--24, 2021, Virtual Event, China}
\acmPrice{15.00}
\acmDOI{10.1145/3474085.3475397}
\acmISBN{978-1-4503-8651-7/21/10}

\acmConference[MM '21]{MM '21: ACM Symposium on Neural
  Gaze Detection}{October 20-21, 2021}{Chengdu, China}
\acmPrice{15.00}

\acmSubmissionID{1343}

\usepackage{graphicx}
\usepackage{subfigure}
\usepackage{balance}

\begin{document}
\fancyhead{}

\title{TransRefer3D: Entity-and-Relation Aware Transformer \\ for Fine-Grained 3D Visual Grounding}

%
\author{Dailan He}
\authornote{Equal contribution.}
\email{hedailan@buaa.edu.cn}
\orcid{1234-5678-9012}
\affiliation{%
  \institution{School of Computer Science and Engineering, Beihang University}
    \city{Beijing}
    \country{China}
}

\author{Yusheng Zhao}
\authornotemark[1]
\email{zhaoyusheng@buaa.edu.cn}
\affiliation{%
  \institution{School of Computer Science and Engineering, Beihang University}
    \city{Beijing}
    \country{China}
}

\author{Junyu Luo}
\authornotemark[1]
\email{luojunyu@buaa.edu.cn}
\affiliation{%
  \institution{School of Computer Science and Engineering, Beihang University}
    \city{Beijing}
    \country{China}
}

\author{Tianrui Hui}
\email{huitianrui@iie.ac.cn}
\affiliation{%
  \institution{Institute of Information Engineering, Chinese Academy of Sciences}
        \city{Beijing}
    \country{China}
}

\author{Shaofei Huang}
\email{huangshaofei@iie.ac.cn}
\affiliation{%
  \institution{Institute of Information Engineering, Chinese Academy of Sciences}
      \city{Beijing}
    \country{China}
}

\author{Aixi Zhang}
\email{aixi.zhax@alibaba-inc.com}
\affiliation{%
 \institution{Alibaba Group}
   \city{Beijing}
 \country{China}
 }

\author{Si Liu}
\email{liusi@buaa.edu.cn}
\authornote{Corresponding author.}
\affiliation{%
  \institution{Institute of Artificial Intelligence}
  \city{Beijing}
  \country{China}
}


\renewcommand{\shortauthors}{He and Zhao, et al.}

\begin{abstract}
Recently proposed fine-grained 3D visual grounding is an essential and challenging task, whose goal is to identify the 3D object referred by a natural language sentence from other distractive objects of the same category.
Existing works usually adopt dynamic graph networks to indirectly model the intra/inter-modal interactions, making the model difficult to distinguish the referred object from distractors due to the monolithic representations of visual and linguistic contents. 
In this work, we exploit Transformer for its natural suitability on permutation-invariant 3D point clouds data and propose a \textit{TransRefer3D} network to extract entity-and-relation aware multimodal context among objects for more discriminative feature learning. 
Concretely, we devise an Entity-aware Attention (EA) module and a Relation-aware Attention (RA) module to conduct fine-grained cross-modal feature matching. 
Facilitated by co-attention operation, our EA module matches visual entity features with linguistic entity features while RA module matches pair-wise visual relation features with linguistic relation features, respectively. 
We further integrate EA and RA modules into an Entity-and-Relation aware Contextual Block (ERCB) and stack several ERCBs to form our TransRefer3D for hierarchical multimodal context modeling. 
Extensive experiments on both Nr3D and Sr3D datasets demonstrate that our proposed model significantly outperforms existing approaches by up to \textbf{10.6\%} and claims the new state-of-the-art performance. 
To the best of our knowledge, this is the first work investigating Transformer architecture for fine-grained 3D visual grounding task.
\end{abstract}





\maketitle


\section{Introduction}

\begin{figure*}
    \centering
    \includegraphics[width=\textwidth]{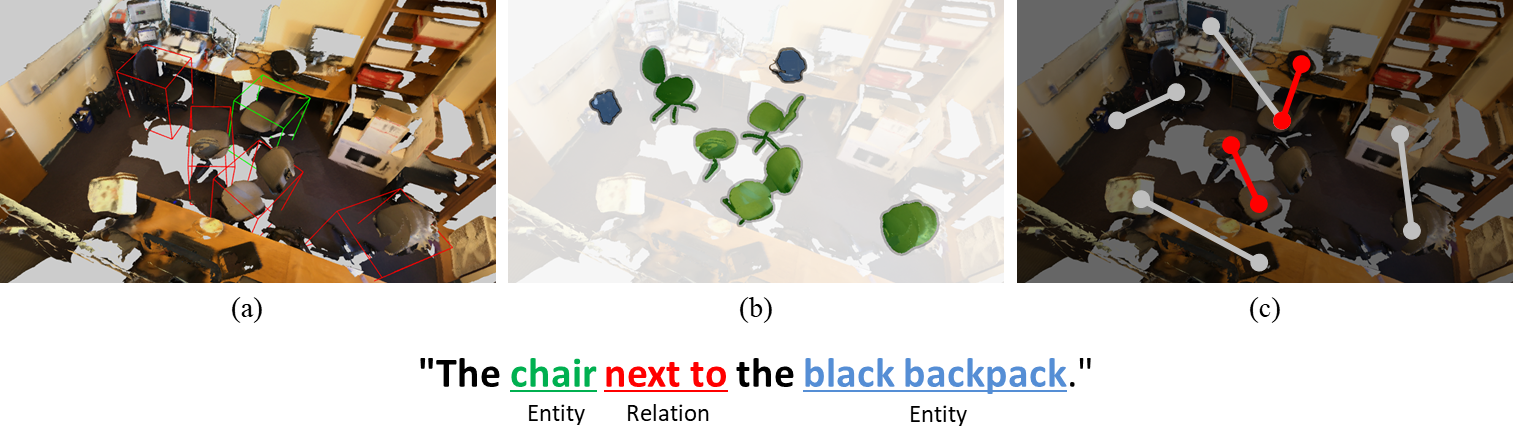}
    \caption{An example of 3D visual grounding (a). According to the input sentence (referring expression) the model is required to locate the referred object (the chair in the given case) from the 3D scene. To perform that, the model needs to identify the \textit{entity} (b) and \textit{relation} (c) mentioned in the referring expression and match them with the corresponding visual elements to obtain a multimodal context in an explicit or implicit way.}
    \label{fig:intro}
\end{figure*}

Visual-linguistic modeling, which aims to connect vision and language, is an essential task in multimedia understanding and embodied AI. Specific tasks like visual question answering~\cite{lu2016hierarchical,yu2018beyond,yu2019deep}, image captioning~\cite{anderson2018bottom,huang2019attention,cornia2020meshed}, image-text matching~\cite{li2019visual,liu2020graph,wang2020consensus}, visual relation detection~\cite{chiou2021visual} and scene graph generation~\cite{zellers2018neural} have been widely studied in the past few years.
In this paper, we focus on another important task named visual grounding, which aims to locate referred objects (referents) given the input referring expressions.
However, most of the existing works on visual grounding have been conducted on 2D images, which might fail to capture the full information in our 3D reality.

Recently, approaches and benchmarks for 3D visual-linguistic tasks emerge and attract attention~\cite{qi2017pointnet,qi2017pointnet++,phan2018dgcnn,ding2019votenet}. 
As an essential 3D visual-linguistic task, 3D visual grounding is to locate objects in 3D point cloud scenes. Two important benchmarks, ScanRefer~\cite{chen2019scanrefer} and ReferIt3D~\cite{achlioptas2020referit3d}, have been proposed based on ScanNet dataset~\cite{dai2017scannet}. While the goal of ScanRefer is to predict 3D bounding boxes for the referents given natural language description, ReferIt3D are more focused on the following two aspects:
(1) The objects in the scenes are annotated with \textbf{fine-grained} labels and referring expressions. For instance, given the description "choose the armchair in the center of the room", the model is required to locate only armchairs but not other types of chairs like folded chairs or office-chairs;
(2) Each scene contains \textbf{multiple} objects of the same fine-grained category as the referent. As a result, the model cannot simply rely on object classification.
Therefore, ReferIt3D requires more fine-grained reasoning of 3D objects and referring expressions, which is more challenging than ScanRefer. In this work, we mainly focus on this fine-grained visual grounding benchmark.

Based on the above benchmarks, researchers have proposed a few 3D visual grounding methods.
ReferIt3DNet~\cite{achlioptas2020referit3d} adopts a dynamic graph network (DGCNN~\cite{phan2018dgcnn}) to model the multimodal context. InstanceRefer~\cite{yuan2021instancerefer} utilizes three parallel modules to match visual and linguistic features from different aspects. However, all of these methods fall short of explicit modeling of entities and their relationships in the cross-modality context. Moreover, some of the previous methods adopt simple fusion of visual feature and linguistic feature. Consequently, the performance of these methods is relatively weak on ReferIt3D benchmark.

We have two observations with a deeper analysis of previous works and the 3D visual grounding task.
The first is \textbf{the similarity between point cloud encoders and Transformers}~\cite{vaswani2017attentionisallyouneed, zhao2020point}. In the 3D visual grounding task, the scene is represented by a set of point cloud features. 
These features are \textit{permutation-invariant}, and preserving this property has been a basic guidance of designing point cloud encoders. 
On the other side, the multi-head self-attention in Transformers preserves this property by its nature. Since Transformers have achieved phenomenal success in both natural language processing~\cite{devlin2018bert} and computer vision~\cite{dosovitskiy2020vit}, we are highly motivated to introduce Transformers in our task. 

Another observation is the importance of \textbf{combining entity and relation information} in visual-linguistic modeling. \textit{Entity} information refers to the feature within an instance, like the category, color, or size. \textit{Relation} information refers to the feature among objects, including spatial relation, comparative relation, etc. Consider the scene "The chair next to the black backpack" (shown in Figure~\ref{fig:intro}), "the chair" and "the black backpack" are linguistic entities, and the corresponding objects are visual entities. Similarly, "next to" is a linguistic relation, and the corresponding spatial relation is implied in the scene. As we can see, the entity or relation information alone is not enough for modeling the multimodal context. Therefore, combining two types of information is essential in our 3D visual grounding task.

Based on these two observations, we propose a Transformer architecture named \textit{TransRefer3D} to better comprehend the entity-and-relation multimodal context for fine-grained 3D visual grounding. Concretely, we utilize Self-Attention modules (SA) to capture intra-modal contextual information. For inter-modal context, we propose an \textit{Entity-aware Attention} (EA) module and a \textit{Relation-aware Attention} (RA) module to match entity and relation features in both modalities. The EA module performs matching between object features and linguistic entity features via co-attention. The RA module first extracts visual relation features of object-object pairs via an asymmetric operator. It then performs cross-modal attention between visual relation features and linguistic relation features. With SA, EA and RA, we integrate them into a unified module called Entity-and-Relation aware Contextual Block (ERCB) for multimodal context modeling. Finally, several ERCBs are stacked together to form our TransRefer3D model that captures the hierarchical feature in the cross-modality context.

To summarize, we contribute to the research on 3D visual grounding from three aspects:
\begin{itemize}
    \item  To the best of our knowledge,  we are the first to introduce the Transformer architecture to achieve a better cross-modal feature representation for the 3D visual grounding task.
    \item We propose entity-and-relation aware attention for multimodal context comprehension and use it for fine-grained 3D visual grounding task.
    \item The proposed TransRefer3D model achieves state-of-the-art performance on ReferIt3D. Compared with existing cutting-edge approaches, the test accuracy improves up to $10.4\%$ on Nr3D (with Sr3D+ augmentation).
\end{itemize}

\section{Related Works}

\begin{figure*}
    \centering
    \includegraphics[width=17cm]{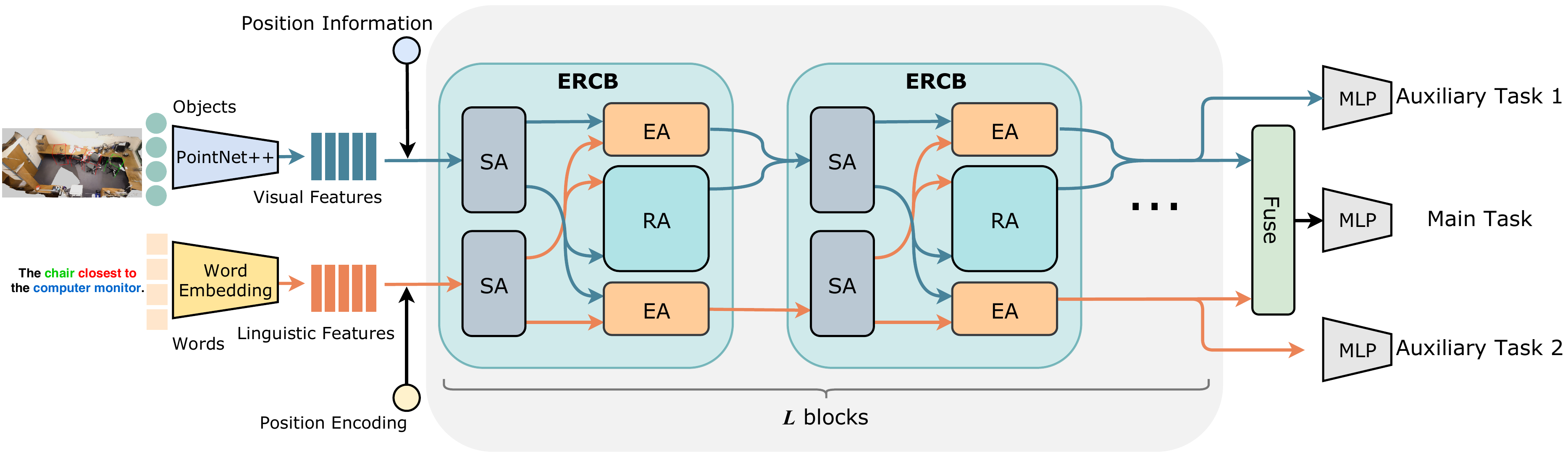}
    \caption{Architecture of the proposed TransRefer3D. The model is composed of several Entity-and-Relation aware Contextual Blocks (ERCBs) stacked together. Each block contains self-attention (SA), Entity-aware Attention (EA) and Relation-aware Attention (RA). The model predicts the referred object (Main Task) as well as an utterance classification of the referent (Auxiliary Task 1) and an object classification (Auxiliary Task 2) for better feature extraction. The blue arrows indicates the flow of visual features, while the red arrows showcases the flow of linguistic features.}
    \label{fig:arch}
\end{figure*}

\subsection{2D Visual Grounding}

Visual grounding, also known as phrase grounding, aims to localize descriptive words in images. It has been widely studied in the past several years\cite{yu2020cross,yang2019fast,zhang2019referring,jing2020visual}. Karpathy \textit{et al.}~\cite{karpathy2014deep} proposed to map image fragments and sentence fragments into a common space to match images and sentences. Chen \textit{et al.}~\cite{chen2017query} proposed a Query-guided Regression Network with Context policy (QRC net) and introduce reinforcement learning techniques to achieve better localization. Huang \textit{et al.}~\cite{huang2020referring}, proposed a Cross-Modal Progressive Comprehension module (CMPC) and a Text-Guided Feature Exchange module (TGFE) to segment referred entities in the image. 
Though researchers have made significant progress on 2D visual grounding, the research on 3D visual grounding is still relatively preliminary. In this work, we mainly focus on fine-grained 3D visual grounding task.

\subsection{3D Visual Grounding}

3D visual grounding is an emerging research topic. Chen~\textit{et al.}~\cite{chen2019scanrefer} released a new benchmark of 3D object localization with natural language descriptions. They proposed ScanRefer to combine visual detection and language encoder for joint inference. Achlioptas~\textit{et al.}~\cite{achlioptas2020referit3d} released another benchmark called ReferIt3D consisting of Nr3D, Sr3D and Sr3D+ datasets, which assumes the segmented object instances are already given. However, its task is more fine-grained where each referred object has distractors of the same category. Text-guided Graph Neural Network (TGNN~\cite{huang2021text}) are proposed for referring instance segmentation on point cloud. Yuan~\textit{et al.}~\cite{yuan2021instancerefer} proposed InstanceRefer which utilizes instance attribute, relation and localization perceptions for 3d visual grounding.
Different from previous works, we exploit the Transformer architecture and propose Entity-aware Attention (EA) module and Relation-aware Attention (RA) module to build the basic blocks of Transformer to achieve a finer matching between visual and linguistic features.

\subsection{Attention Mechanism and Transformer}

Transformers have been widely used in natural language processing ~\cite{vaswani2017attentionisallyouneed, devlin2018bert, liu2019roberta}. Recently a lot of works also apply them on high-level vision tasks ~\cite{dosovitskiy2020vit, carion2020detr}, low-level vision tasks ~\cite{chen2020ipt} and graph tasks~\cite{yun2019graph} because of its strong ability to recognize long-term dependency.

A Transformer is composed of several Transformer layers stacked together. Each Transformer layer consists of a multi-head self-attention (MSA) and a feed forward network. MSA is based on the self-attention mechanism. Given three feature matrices: queries $Q \in \mathbb{R}^{n \times d}$, keys and values $K, V \in \mathbb{R}^{m\times d}$, a single-head self-attention is formulated as:
\begin{equation}
    \mathrm{Attn}(Q, K, V) = \mathrm{softmax}\left(\frac {QK^\top}{\sqrt{d}} \right) V
\end{equation}
introducing $H$ parallel attention layers as different transformation heads to enlarge the model capacity, we can further obtain a multi-head attention:

\begin{equation}
\begin{aligned}
    \mathrm{MSA}(Q, K, V) = W_{out}\left[f_1, f_2, \dots, f_H \right] \\
    f_i = \mathrm{Attn}\left(W_q^{(i)}Q, W_k^{(i)}K, W_v^{(i)}V\right)
\end{aligned}
\end{equation}
where $W_q^{(i)}, W_k^{(i)}, W_v^{(i)} \in \mathbb{R}^{d_h\times d}$ are linear transformation that projects $Q,K,V$ for the $i$-th head and $W_{out} \in \mathbb{R}^{h \times Hd_h}$ for the output. After the multi-head attention follows a feed forward network, which is a multi-layer perceptron (MLP). Moreover, the Transformer also utilize Layer Normalization and residual connections.

\subsection{Co-Attention for Multimodal Learning}

Multimodal learning requires a full understanding of the contents in each modality and, more importantly, learning the interactions between modalities. Co-attention mechanism~\cite{lu2016hierarchical} is proposed to modeling the image attention and question attention in VQA. Yu \textit{et al.}~\cite{yu2019deep} proposed a dense co-attention model to make full interactions between modalities and address the previous issue that each modality learns separate attention distributions. In 3D visual grounding tasks, the context of entity-and-relation has not been fully researched yet. 
In this paper, we expand the co-attention mechanism to guide context modeling in 3D visual grounding task. This is the first work investigating the co-attention mechanism on the 3D visual-linguistic tasks to the best of our knowledge. 

\section{Methodology}

\subsection{Overview}

According to the above sections, scenes contain two types of information: entities' attributes and relations among entities. 
To better model the multimodal context for fine-grained 3D visual grounding, we propose two cross-modal attention mechanism: Entity-aware Attention (EA) and Relation-aware Attention (RA). Furthermore, we develop an Entity-and-Relation aware Contextual Block (ERCB) that integrate those attention mechanisms as a multimodal encoding unit. By stacking multiple ERCBs, we obtain the proposed TransRefer3D model.

Figure~\ref{fig:arch} shows the framework of our model. Following previous work ~\cite{achlioptas2020referit3d}, we first extract object features with a shared PointNet++~\cite{qi2017pointnet++} and vectorize the input words with an embedding layer. After that, we fuse position information to both visual and linguistic features and feed them into the proposed TransRefer3D.

We use the same training objective as ReferIt3D ~\cite{achlioptas2020referit3d} consisting of a main task and two auxiliary tasks. For $N$ input objects, the main training task is a $N$-classification supervised by a cross-entropy loss $\mathcal{L}_{\mathrm{main}}$. To obtain the features well embedded to represent $K$ categories, we also adopt $K$-classification on object features and aggregated linguistic features, supervised by cross entropy loss $\mathcal{L}_\mathrm{obj}$ and $\mathcal{L}_\mathrm{lang}$. The total loss function is $\mathcal{L} = \mathcal{L}_\mathrm{main} + \lambda_\mathrm{obj} \mathcal{L}_\mathrm{obj} + \lambda_\mathrm{lang} \mathcal{L}_\mathrm{lang}$ where we set $\lambda_\mathrm{obj}=\lambda_\mathrm{lang}=0.5$ in all of our experiments.

\subsection{\textbf{EA:} Entity-aware Attention Module}

\begin{figure}
    \centering
    \includegraphics[width=4.5cm]{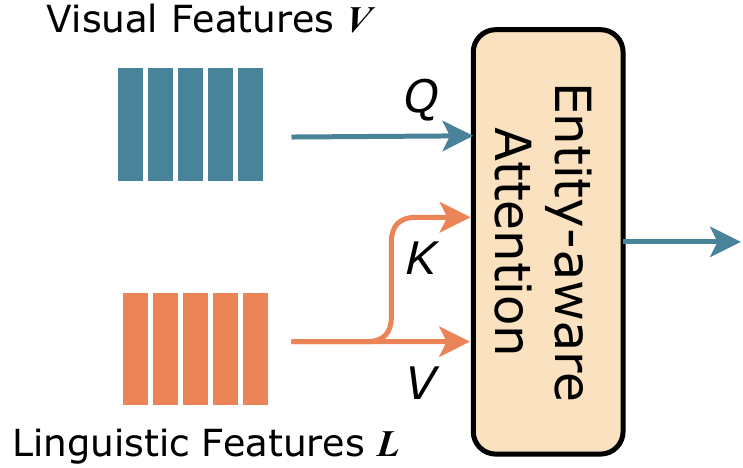}
    \caption{
    The Entity-aware Attention module (EA). The figure illustrates the language-guided visual entity-aware attention, where visual features serve as queries and linguistic features are keys and values of multi-head attention. The vision-guided linguistic entity-aware attention is similar, with linguistic features as the queries and visual features as keys and values.
    }
    \label{fig:EA}
\end{figure}

Our goal is to find out the referent according to the referring expression. Usually, the referent and other objects in the scene get mentioned in the referring expression as some keywords. We name all objects appearing in the scene and their corresponding keywords as the \textbf{\textit{entities}} in each modality. First of all, we establish an Entity-aware Attention (EA) module for finely understanding the multimodal context.

The importance of each entity for locating the referent is different. In a ScanNet indoor scene, often there are more than 20 objects, but most of them make a limited contribution to the visual grounding task. To better understand the referring expression, a strategy is adapt to match the entity objects with corresponding words and suppress relatively useless features. By this, the model can finely identify and establish correlations on those important entities.

Inspired by self-attention mechanism and co-attention widely adopted in 2D vision-language approaches~\cite{huang2020referring, yu2019deep}, we propose EA module to match those features to model an entity aware multimodal context. In a 3D scene, the $n$ objects are represented by $d$-dimension features. And the referring expression is handled to $m$ word-level features. 
Given two feature sets in different modality $X \in \mathbb{R}^{n\times d}$ and $Y \in \mathbb{R}^{m \times d}$, the EA module matches $Y$ to $X$:
\begin{equation}
    \mathrm{EA}(X, Y)  = \mathrm{Attn}(X, Y, Y) = \mathrm{softmax}\left(\frac {X Y^\top} {\sqrt{d}}\right) Y
\end{equation}
which is a multimodal generalization of attention. 
Like self-attention, we further extend EA to a multi-head form by replacing $Attn(\cdot)$ with a multi-head version. 

The proposed EA module does symmetrical matchings to visual and word-level linguistic features for better modeling entities in both modals.

\subsection{\textbf{RA:} Relation-aware Attention Module}
\label{sec:method-ra}

The EA module can match entity information in word features and 3D object features. However, only focusing on the entities is insufficient for complicated visual-linguistic reasoning. For example, consider the scene with multiple objects of the same category as the referent. All of them match the corresponding word feature, so their corresponding Entity-aware Attention scores can be similar. Consequently, it can be challenging for the model to distinguish the referent from others.

To address this problem, it is important to let model recognize those relational words (e.g. "next to") that appear in the referring expression and enhance entity features according to their relations. We establish a representation of abstract entity-to-entity relation for visual modality and develop a cross-modal Relation-aware Attention (RA) for finer context modeling.

Inspired by DGCNN, we define the relation representation between two entity features $f_i, f_j$ as an asymmetric form:
\begin{equation}
    r_{ij} = r(f_i, f_j) = H_{\Theta}(f_i-f_j)
    \label{eq:relation-representation}
\end{equation}
where $H_\Theta$ denotes a nonlinear layer.

\begin{figure}
    \centering
    \includegraphics[width=6cm]{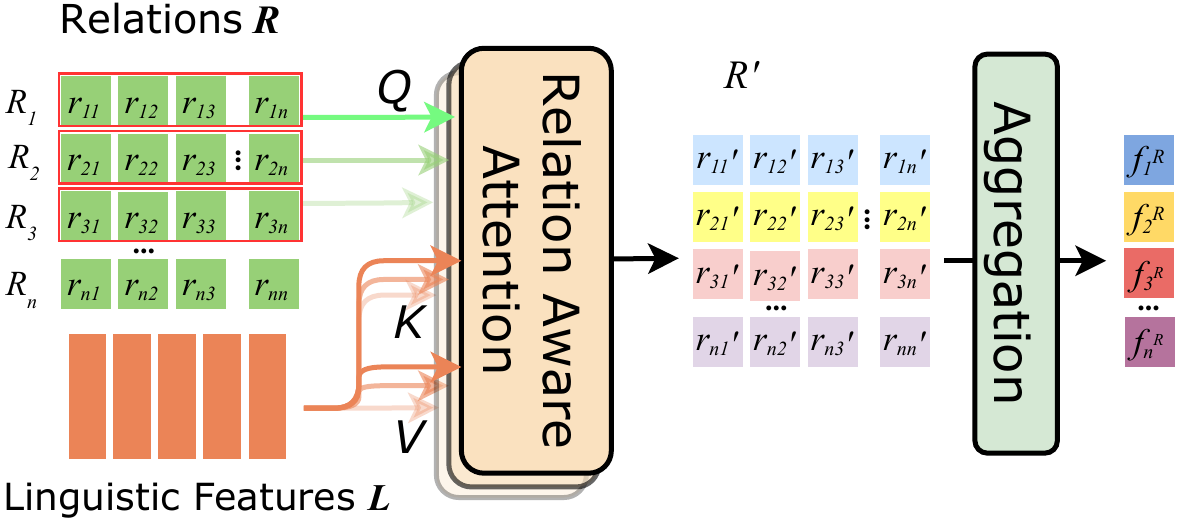}
    \caption{
    The Relation-aware Attention module (RA). In RA, the visual features are first extracted and then matched with the language feature. The relation features serve as queries, and the linguistic features are keys and values. After the cross-modal attention, the language-guided relation features are then aggregated.
    }
    \label{fig:RA}
\end{figure}

Like entity features, each relation features $r_{ij} \in \mathbb{R}^{d}$ can also get matched with corresponding linguistic features. For each entity feature $f_i$, we calculate its relation features and introduce co-attention on them to obtain an enhanced cross-modal context of $f_i$:
\begin{equation}
\begin{aligned}
    R_i &= [r_{i1}, r_{i2}, \dots, r_{in}] \\
    R'_i &= \mathrm{Attn}(R_i, L, L) \\
    f^R_i &= \mathrm{Agg}( R'_i ) \\
    F^R_i &= [f^R_1, f^R_2, \dots, f^R_n]
\end{aligned}
\label{eq:RA}
\end{equation}
where $L$ is the linguistic feature vector and $\mathrm{Agg}(\cdot)$ denotes a channel-wise aggregation operator gathering the $n$ fused $d$-dimension features in $R'_i$ to $f^R_i \in \mathbb{R}^d$. Figure~\ref{fig:RA} shows the diagram of RA. Hence for each $f_i$ we can obtain a corresponding $f^R_i$ as its relation-aware multimodal context representation.

We adopt a binary relation representation calculated on each pair of entities (eq.~\ref{eq:relation-representation}). Assuming that those more complicated relations among multiple entities can be described as a set of binary relations, we stack RA modules in our model to achieve a hierarchical and progressive context comprehension with respect to relation information.

\begin{figure}
    \centering
    \includegraphics[width=7cm]{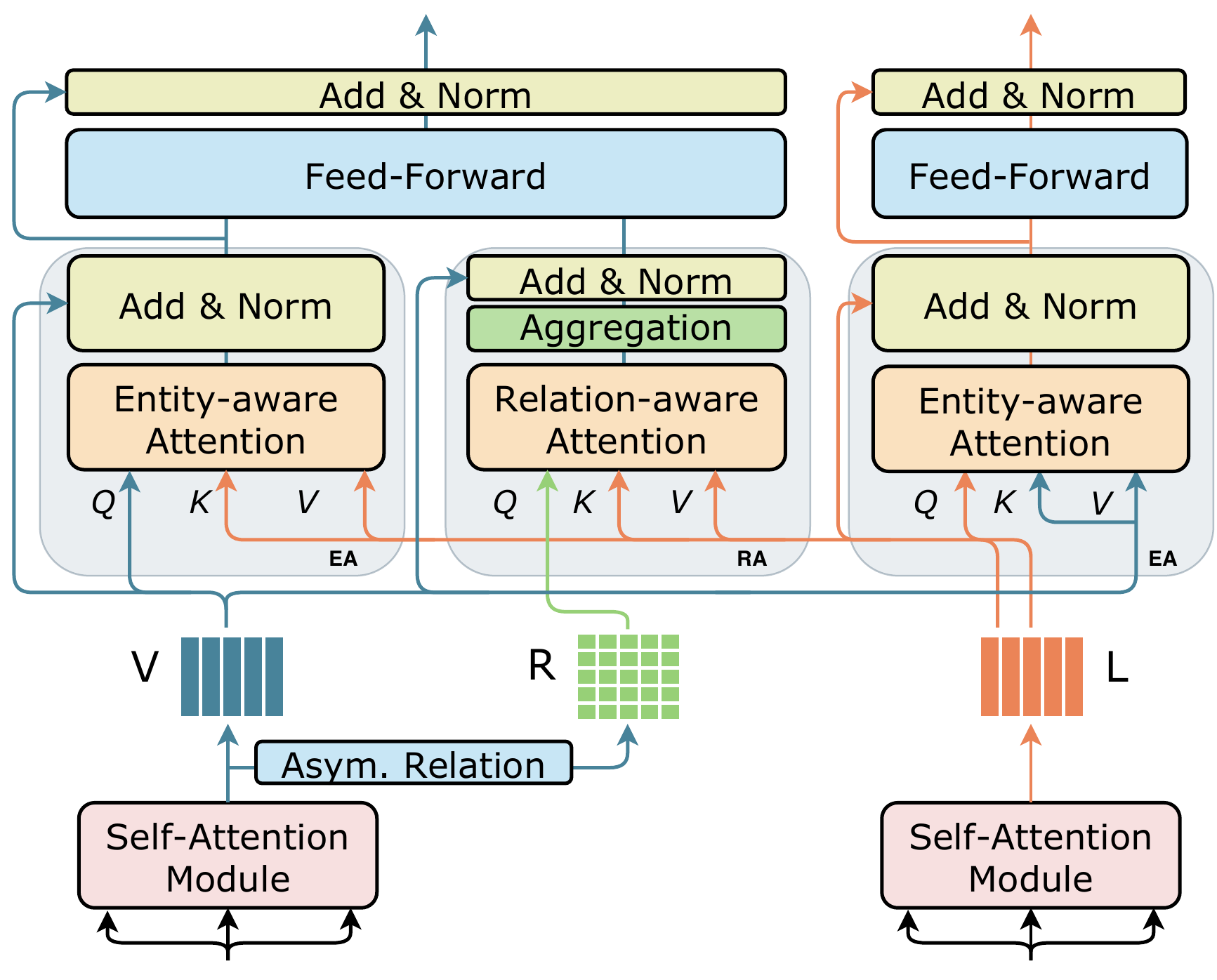}
    \caption{The architecture of the proposed ERCB. The input visual and linguistic features are first fed into two self-attention modules. Then visual entity features and linguistic entity features are matched through Entity-aware Attention (EA). The visual relation features are first extracted and then matched with linguistic features through Relation-aware Attention. Finally, the enhanced features are fused and processed by feed-forward networks.}
    \label{fig:block}
\end{figure}

\subsection{\textbf{ERCB:} Entity-and-Relation Aware Contextual Block}
EA and RA can model two essential information in scenes: entities and relations. By composing EA and RA, we propose an Entity-and-relation Aware Contextual Block (ERCB), as shown in Figure~\ref{fig:block}. ERCB can form a cross-modal context comprehension with respect to both entity and relation information for visual grounding.

In each block, we first feed the visual and linguistic features to self-attention layers. Self-attention layers can effectively extract features in each modality and enhance the performance of cross-modal transformations.

Then, we match the visual features with word-level features with respect to both entity information and relation information. We calculate two types of attention representation $F^E$ and $F^R$ in parallel and fuse them to obtain the joint features $f^{J}$:
\begin{equation}
\begin{aligned}
    g^{E}_i = \mathrm{LN}(f^E_i+f_i) \\
    g^{R}_i = \mathrm{LN}(f^R_i+f_i) \\ 
    f^{J}_i = \mathrm{FFN}([g^{E}_i, g^{R}_i ]) \\
\end{aligned}
\end{equation}
where $f^J_i \in \mathbb{R}^d$. $f^E_i$ and $f^R_i$ are the $i$-th feature of $F^E$ and $F^R$, respectively.  $\mathrm{FFN}$ is a feed-forward network and $\mathrm{LN}$ denotes layer normalization~\cite{ba2016layer}. Further performing an add-and-norm transformation, we can obtain the enhanced multimodal feature $f'_i$:
\begin{equation}
    f'_i = \mathrm{LN}(f^{J}_i + g^{E}_i)
\end{equation}

We also enhance linguistic features with EA guided by visual features. For each attention layer, a feed-forward network is followed to introduce non-linearity. By stacking those ERCBs, we finally obtain a hierarchical Transformer for fine-grained multimodal context comprehension.

In Section~\ref{sec:abl}, we will discuss the ability of each module and the structure of our model. Experiments show that TransRefer3D and its component EACB are much more powerful in modeling multimodal context than other methods.

\section{Experiments}

\begin{table*}[th]
    \centering
    \begin{tabular}{c||c||c|c|c|c}
    \hline
Arch. &  Overall & Easy & Hard & View-dep. & View-indep. \\
 \hline \hline
   \multicolumn{6}{c}{\textbf{Nr3D}} \\
\hline

ReferIt3DNet~\cite{achlioptas2020referit3d} & $35.6\% \pm 0.7\%$ & $43.6\% \pm 0.8\%$ & $27.9\% \pm 0.7\%$ & $32.5\% \pm 0.7\%$ & $37.1\% \pm 0.8\%$\\
TGNN~\cite{huang2021text} & $37.3\% \pm 0.3\%$ & $44.2\% \pm 0.4\%$ & $30.6\% \pm 0.2\%$ & $35.8\% \pm 0.2\%$ & $38.0\% \pm 0.3\%$ \\
InstanceRefer~\cite{yuan2021instancerefer} & $38.8\% \pm 0.4\%$ & $46.0\% \pm 0.5\%$ & $31.8\% \pm 0.4\%$ & $34.5\% \pm 0.6\%$ & $41.9\% \pm 0.4\%$\\
TransRefer3D (ours) 
&$\mathbf{42.1\% \pm 0.2\%}$ &$\mathbf{48.5\% \pm 0.2\%}$ &$\mathbf{36.0\% \pm 0.4\%}$ &$\mathbf{36.5\% \pm 0.6\%}$
&$\mathbf{44.9\% \pm 0.3\%}$\\
 
\hline\hline
\multicolumn{6}{c}{\textbf{Nr3D} w/ \textbf{Sr3D}} \\
\hline
ReferIt3DNet & $37.2\% \pm 0.3\%$ & $44.0\% \pm 0.6\%$ & $30.6\% \pm 0.3\%$ & $33.3\% \pm 0.6\%$& $39.1\% \pm 0.2\%$ \\
TransRefer3D (ours) 
& $\mathbf{47.2\% \pm 0.3\%}$ 
& $\mathbf{55.4\% \pm 0.5\%}$ 
& $\mathbf{39.3\% \pm 0.5\%}$ 
& $\mathbf{40.3\% \pm 0.4\%}$ 
& $\mathbf{50.6\% \pm 0.2\%}$ \\
\hline
\hline
\multicolumn{6}{c}{\textbf{Nr3D} w/ \textbf{Sr3D+}} \\
\hline
ReferIt3DNet & $37.6\% \pm 0.4\%$ & $45.4\% \pm 0.6\%$ & $30.0\% \pm 0.4\%$ & $33.1\% \pm 0.5\%$ & $39.8\% \pm 0.4\%$ \\
TransRefer3D (ours) & $\mathbf{48.0\% \pm 0.2\%}$ &  
$\mathbf{56.7\% \pm 0.4\%}$ & 
$\mathbf{39.6\% \pm 0.2\%}$ &  
$\mathbf{42.5\% \pm 0.2\%}$ & 
$\mathbf{50.7\% \pm 0.4\%}$\\
\hline
\end{tabular}
\caption{The performance of TransRefer3D on Nr3D trained with or without Sr3D/Sr3D+, compared with previous works.}
\label{tab:performance-nr3d}
\end{table*}

In this section, we present experiments to demonstrate the effectiveness of the proposed TransRefer3D compared with other methods. We also conduct quantitative/qualitative studies to analyze our method in details.

\subsection{Implementation details and Datasets}

For a fair comparison, we follow the same settings as ReferIt3D~\cite{achlioptas2020referit3d}. We train the model until convergence with Adam optimizer~\cite{kingma2014adam} and a learning rate of 0.0005. The default feature dimension ($d$) is 128 and the default batch size is 32. We set the depth of ERCBs ($L$) to 4 and use 4-head attentions if not particularly indicated.

We conduct our experiments on three datasets, Nr3D, Sr3D and Sr3D+~\cite{achlioptas2020referit3d}. The details of these datasets are listed as follows:
\begin{itemize}
    \item \textbf{Nr3D}: Nr3D (Natural Reference in 3D) consists of 41.5K human descriptions collected using a referring game (ReferIt Game). It describes objects in 707 ScanNet scenes. In each scene, there are multiple objects of the same category as the referent.
    \item \textbf{Sr3D}: Sr3D (Spatial Reference in 3D) contains 83.5K synthetic descriptions. It categorizes spatial relations into 5 types: horizontal proximity, vertical proximity, between, allocentric and support~\cite{achlioptas2020referit3d}, and then generates descriptions using language templates. Similar with Nr3D, there are always more than one objects of the same category as the referred object.
    \item \textbf{Sr3D+}: Sr3D+ is an augmentation of Sr3D. It uses the same method as Sr3D to generate synthetic descriptions. The difference is the referent could be the only object belonging to a fine-grained category in each scene.
\end{itemize}

Sr3D and Sr3D+ are designed to enhance the performance of the model on Nr3D~\cite{achlioptas2020referit3d}. The model is first jointly trained on both Nr3D training set and Sr3D/Sr3D+ and then evaluated on Nr3D test set. Nevertheless, for the completeness of  experiments, we also train and test our model only on Sr3D.

\subsection{Quantitative Performance on Nr3D and Sr3D/Sr3D+}

\begin{table*}[]
    \centering
    \begin{tabular}{c||c||c|c|c|c}
    \hline
        Arch. &  Overall & Easy & Hard & View-dep. & View-indep. \\
       \hline
       \hline
ReferIt3DNet~\cite{achlioptas2020referit3d} & $40.8\% \pm 0.2\%$ & $44.7\% \pm 0.1\%$ & $31.5\% \pm 0.4\%$ & $39.2\% \pm 1.0\%$ & $40.8\% \pm 0.1\%$\\
TGNN~\cite{huang2021text}  & $45.0\% \pm 0.2\%$ & $48.5\% \pm 0.2\%$ & $36.9\% \pm 0.5\%$ & $45.0\% \pm 1.1\%$ & $45.0\% \pm 0.2\%$ \\
InstanceRefer~\cite{yuan2021instancerefer} & $48.0\% \pm 0.3\%$ & $51.1\% \pm 0.2\%$ & $40.5\% \pm 0.3\%$ & $45.4\% \pm 0.9\%$ & $48.1\% \pm 0.3\%$\\
TransRefer3D (ours) 
&$\mathbf{57.4\% \pm 0.2\%}$ &$\mathbf{60.5\% \pm 0.2\%}$ &$\mathbf{50.2\% \pm 0.2\%}$ &$\mathbf{49.9\% \pm 0.6\%}$
&$\mathbf{57.7\% \pm 0.2\%}$\\
    \hline
    \end{tabular}
    \caption{The performance of TransRefer3D trained and evaluated on Sr3D only, compared with previous works.}
    \label{tab:performance-sr-only}
\end{table*}

\begin{table*}[ht]
    \centering
    \begin{tabular}{c||c||c|c|c|c}
    \hline
        Arch. &  Overall & Easy & Hard & View-dep. & View-indep. \\
       \hline
       \hline
        w/o SA & $38.6\% \pm 0.2\%$ & $45.5\% \pm 0.6\%$ & $32.0\% \pm 0.4\%$ & $33.3\% \pm 0.7\%$ & $41.3\% \pm 0.3\%$\\
        \hline
        w/o EA (V$\to$L) & $41.7\% \pm 0.2\%$ & $48.7\% \pm 0.5\%$ & $34.9\% \pm 0.1\%$ & $37.7\% \pm 0.5\%$ & $43.7\% \pm 0.2\%$ \\
        w/o EA (L$\to$V) & $41.1\% \pm 0.4\%$ & $48.9\% \pm 0.6\%$ & $33.6\% \pm 0.4\%$ & $37.9\% \pm 0.2\%$ & $42.7\% \pm 0.5\%$ \\
        w/o RA & $40.8\% \pm 0.1\%$ & $47.9\% \pm 0.2\%$ & $34.0\% \pm 0.1\%$ & $36.7\% \pm 0.5\%$ & $42.9\% \pm 0.1\%$\\
        \hline
        EA + RA (stacked) & $41.4\% \pm 0.4\%$ & $47.7\% \pm 0.6\%$ & $35.3\% \pm 0.3\%$ & $37.0\% \pm 0.6\%$ & $43.5\% \pm 0.4\%$\\
        \hline
        EA + RA (TransRefer3D)  &$42.1\% \pm 0.2\%$ &$48.5\% \pm 0.2\%$ &$36.0\% \pm 0.4\%$ &$36.5\% \pm 0.6\%$
      &$44.9\% \pm 0.3\%$ \\
    \hline
    \end{tabular}
    \caption{Ablation study of our model. We train and test the proposed models without self-attention (SA) and the proposed modules (vision-guided EA, language guided EA and RA). We also try different module architectures EA and RA modules stacked together or in parallel (TransRefer3D).}
    \label{tab:abl-visual-ga}
\end{table*}

In this section, we report the quantitative performance of our model on Nr3D, Sr3D and Sr3D+.

\subsubsection{Evaluation on Nr3D} \text{}\\
Firstly, we train our model on Nr3D training set and test the model on Nr3D test set. As shown in Table~\ref{tab:performance-nr3d} (upper part), TransRefer3D outperforms previous models by a large margin. Besides, we also measure our performance on different types of scenarios, defined in ~\cite{achlioptas2020referit3d}:
\begin{itemize}
    \item \textbf{Easy vs. Hard}: easy cases are those having only one distractor (the object of the same class as the referent but not the referent itself) and hard cases are those having multiple distractors.
    \item \textbf{View dependency}: view dependent cases (View-dep.) are those whose descriptions ask the viewers to place themselves facing certain objects, and other cases are view independent (View-indep.) .
\end{itemize}
The results show that TransRefer3D performs especially well in hard cases, suggesting that our model can better understand the complicated referring context.

Then, we also train our model on Nr3D jointly with Sr3D/Sr3D+, and test the model on Nr3D test set. The results are shown in Table~\ref{tab:performance-nr3d} (middle part and bottom part). The results indicate that TransRefer3D may perform even better on larger datasets. Compared with training only on Nr3D, the performance of ReferIt3DNet improves by 1.6\% by joint training on Nr3D and Sr3D. However, we find that this improvement enlarges to 5.1\% on TransRefer3D. A similar trend is also observed on our model by joint training with Sr3D+. One possible explanation for this phenomenon is that training Transformers require relatively more data. Recent studies on Transformers designed for other tasks also report similar conclusions~\cite{devlin2018bert, dosovitskiy2020vit}. Therefore, TransRefer3D may have the potential to achieve better performance on visual grounding if a larger dataset is available in the future.

\subsubsection{Evaluation on Sr3D}\text{}\\
We also train and test the proposed model only on Sr3D, and the results are shown in Table~\ref{tab:performance-sr-only}. The results show that our model outperforms all the previous works with more than $\mathbf{9.4}\%$ accuracy improvement. Since the data in Sr3D is generated from language templates that describe spatial relations, it is relatively easy for TransRefer3D that can explicitly model entity and relation in a cross-modality context.

\subsection{Ablation Study}
\label{sec:abl}

\begin{figure}[ht]
    \centering
    \includegraphics[width=5.5cm]{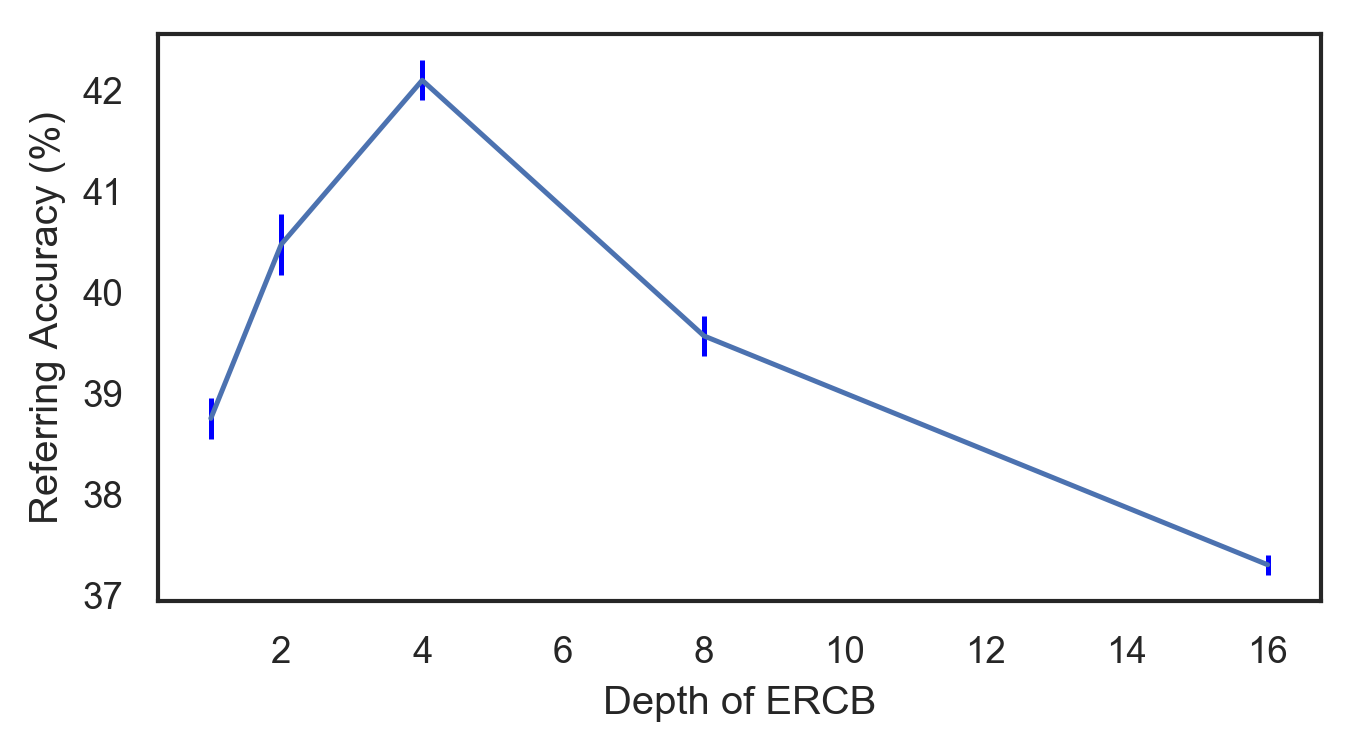}
    \caption{Referring Accuracy with different depth of ERCB.}
    \label{fig:depth-acc}
\end{figure}

\begin{figure}[]
    \centering
        \includegraphics[width=8cm]{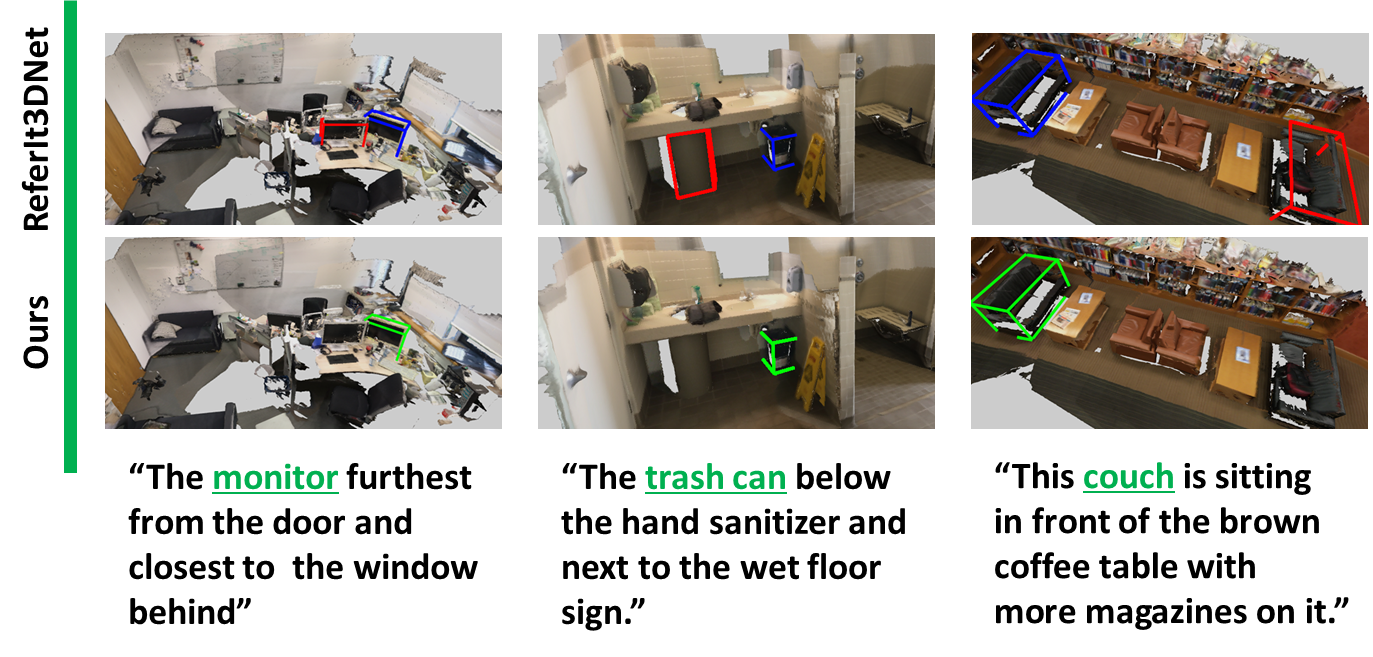}
    \caption{Visualization of the visual grounding results compared with ReferIt3DNet. The blue bounding boxes are the targets, the red ones are misclassifications, and the green ones are the correct classifications.}
    \label{fig:case-study}
\end{figure}

In this subsection, we further discuss the effectiveness of each module in TransRefer3D and the architecture design.

To verify the effectiveness of each module in the proposed model, we train and test TransRefer3D without each of the modules: self-attention (SA), vision-guided linguistic entity aware attention (EA, V$\to$L), language-guided visual entity aware attention (EA, L$\to$V) and relation aware attention (RA). The results are shown in Table~\ref{tab:abl-visual-ga}. The model without SA performs worst in all the experiments. This indicates that the Transformer architecture is suitable for our 3D visual grounding task and that the contextual awareness within a single modal is the basis of cross-modal understanding. Furthermore, removing all of the proposed modules hinders the performance, demonstrating the effectiveness of the proposed EA and RA modules.

Moreover, we train and test our model with different architectures. By default, the proposed ERCB  consists of an EA module and a RA module in parallel. We have also tried stacking EA and RA modules, in which the data is first fed into EA module and then RA module. A possible explanation for this is that the entity information and relation information are relatively independent, and a parallel structure might be more reasonable.

As for the depth of our model, we tested TransRefer3D with different depth of ERCB layers on Nr3D dataset. As shown in Figure~\ref{fig:depth-acc}, 4 layers of ERCB achieve the best performance. Shallow models with fewer layers might fail to represent the complicated relations as analyzed in Section~\ref{sec:method-ra}. Simultaneously, deeper Transformers have been observed hard to train because of unbalanced gradients and amplification effect ~\cite{liu2020understanding}.

\begin{figure*}[htb]
    \subfigure{
        \includegraphics[width=5in]{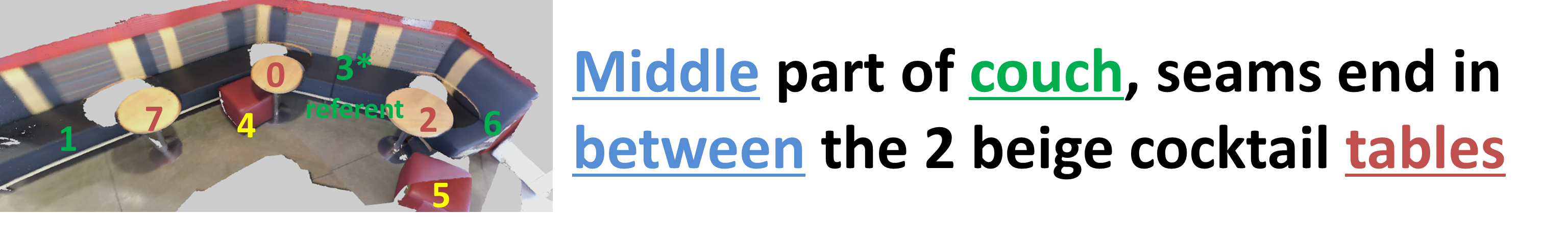}
    }
    \setcounter{subfigure}{0}
    
    \subfigure[EA-1 head 1]{
        \includegraphics[width=4cm]{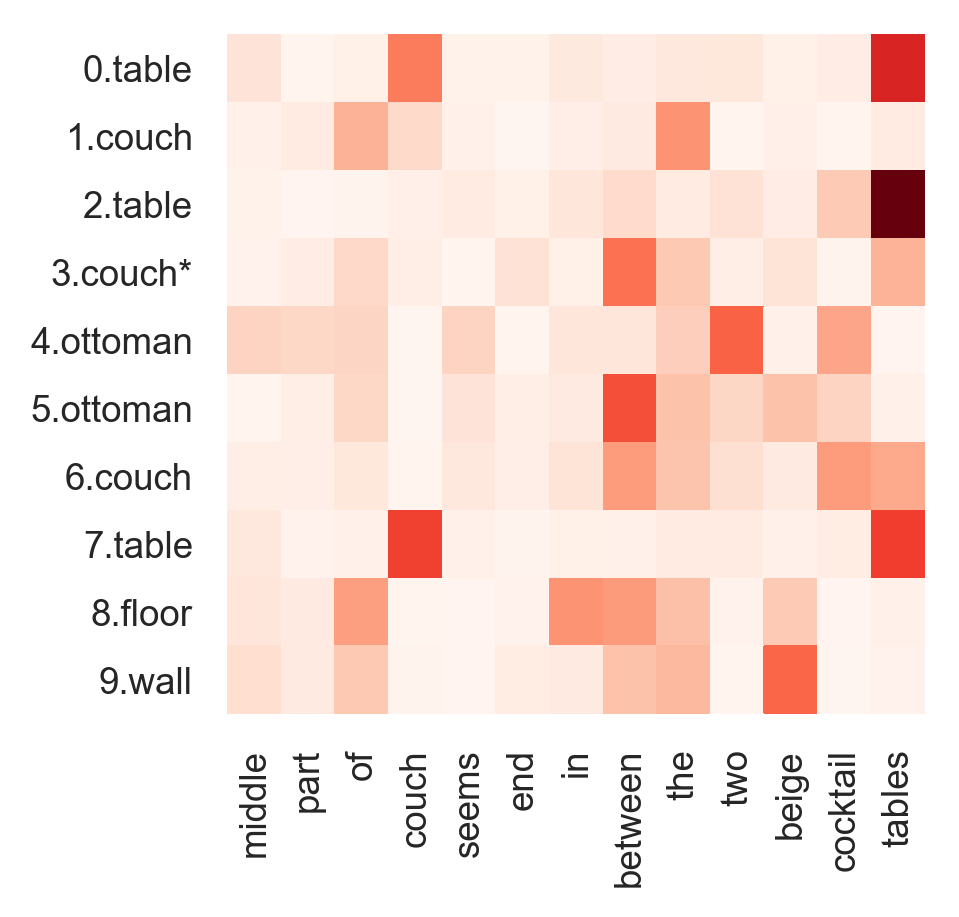}
         \label{fig:EA-1-head-1}
    }
    \subfigure[EA-1 head 2]{
        \includegraphics[width=4cm]{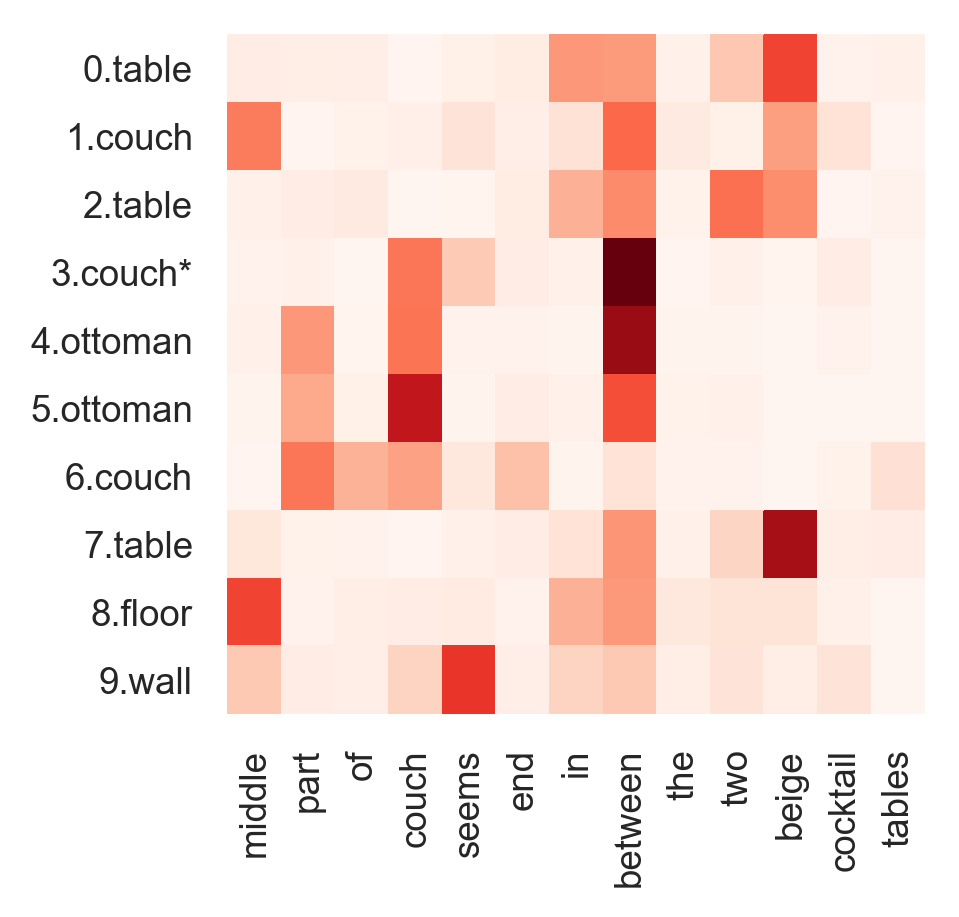}
         \label{fig:EA-1-head-2}
    }
    \subfigure[EA-4 head 1]{
        \includegraphics[width=4cm]{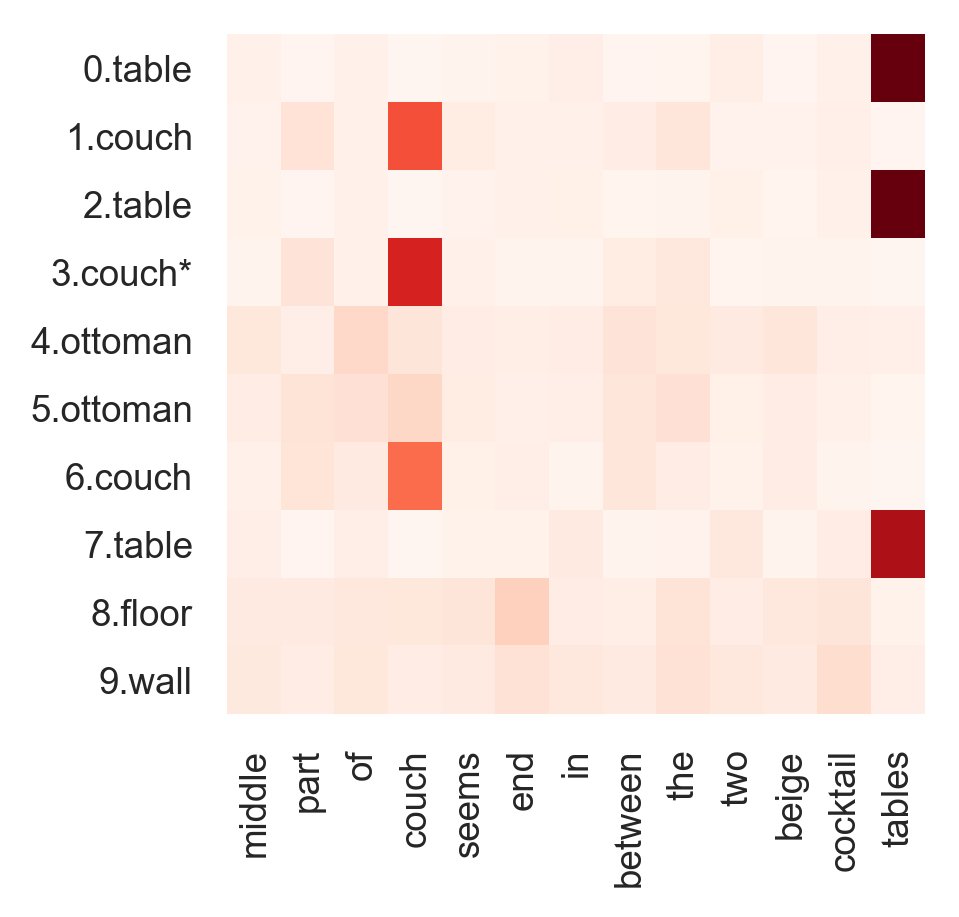}
        \label{fig:EA-4-head-1}
    }
    \subfigure[EA-4 head 2]{
        \includegraphics[width=4cm]{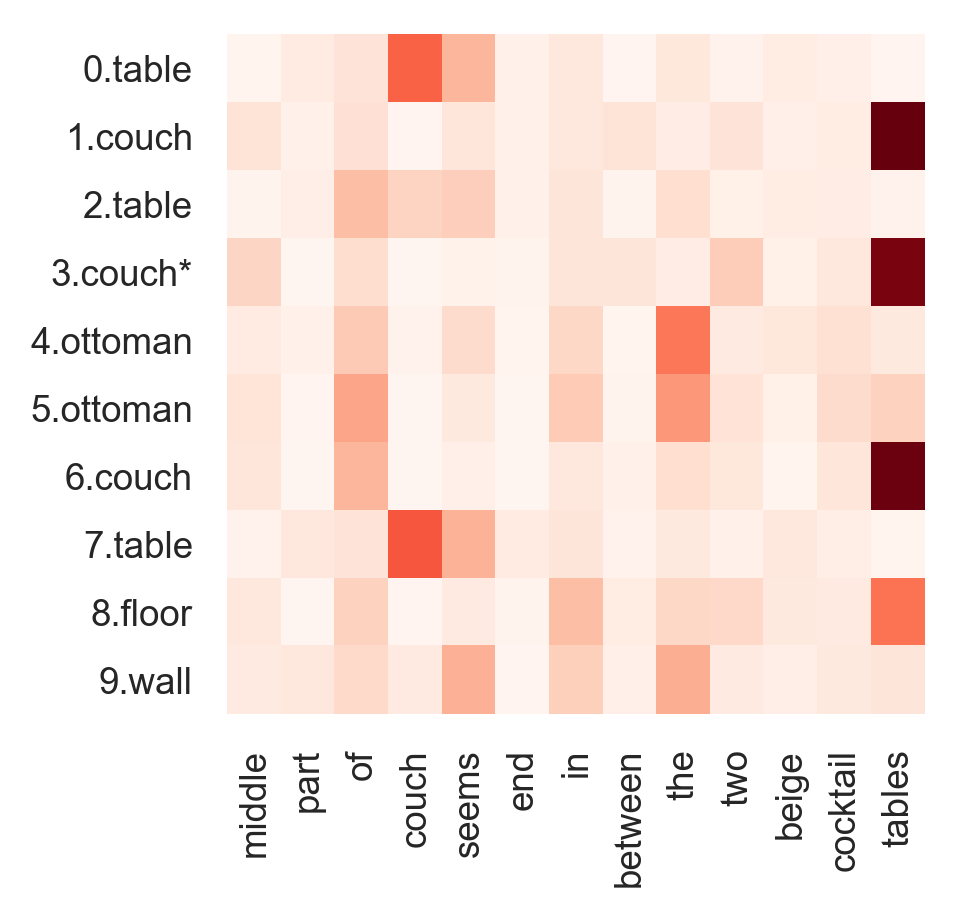}
        \label{fig:EA-4-head-4}
    }
    \subfigure[RA-1 "between"]{
        \includegraphics[width=4cm]{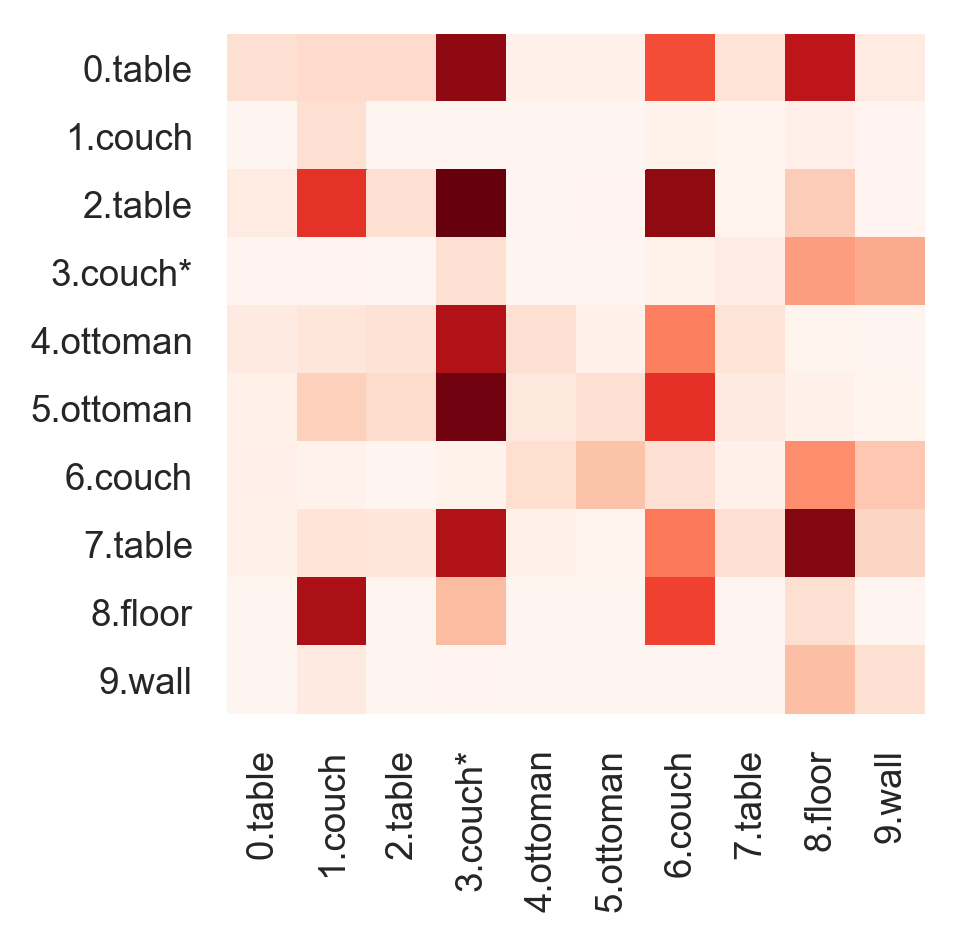}
    }
    \subfigure[RA-4 "between"]{
        \includegraphics[width=4cm]{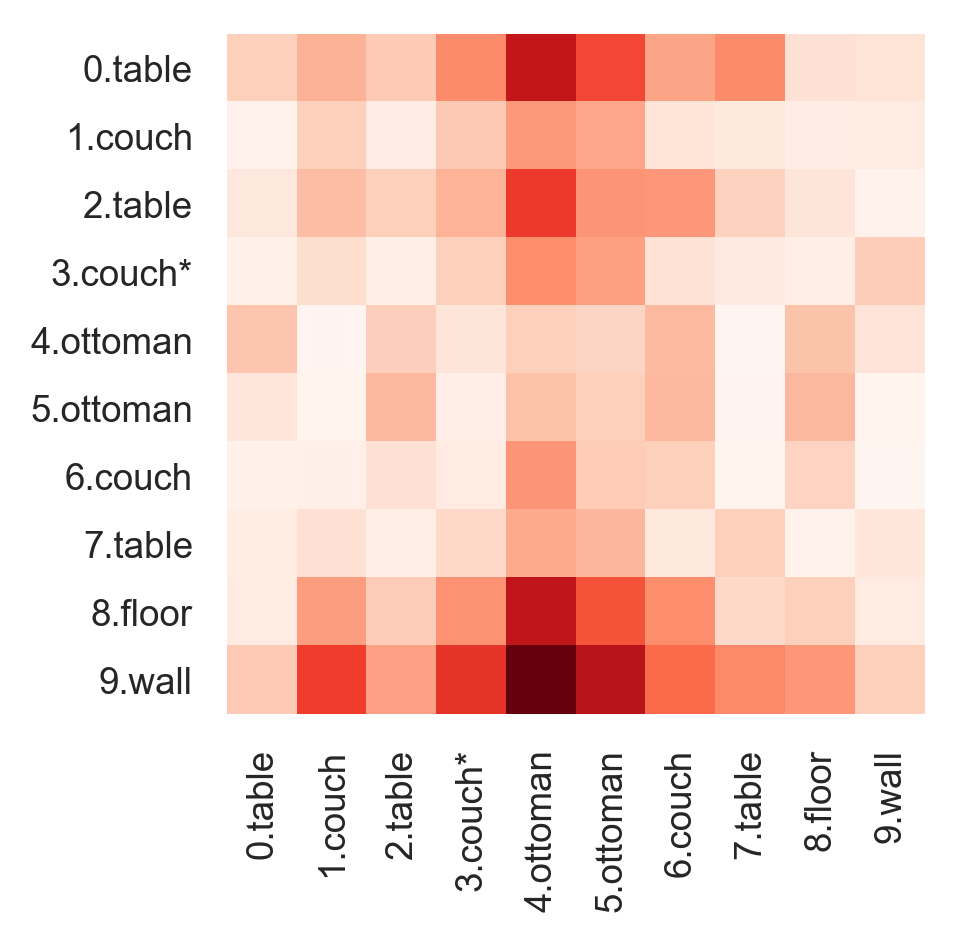}
        \label{fig:RA-4-between}
    }
    \subfigure[RA-1 "middle"]{
        \includegraphics[width=4cm]{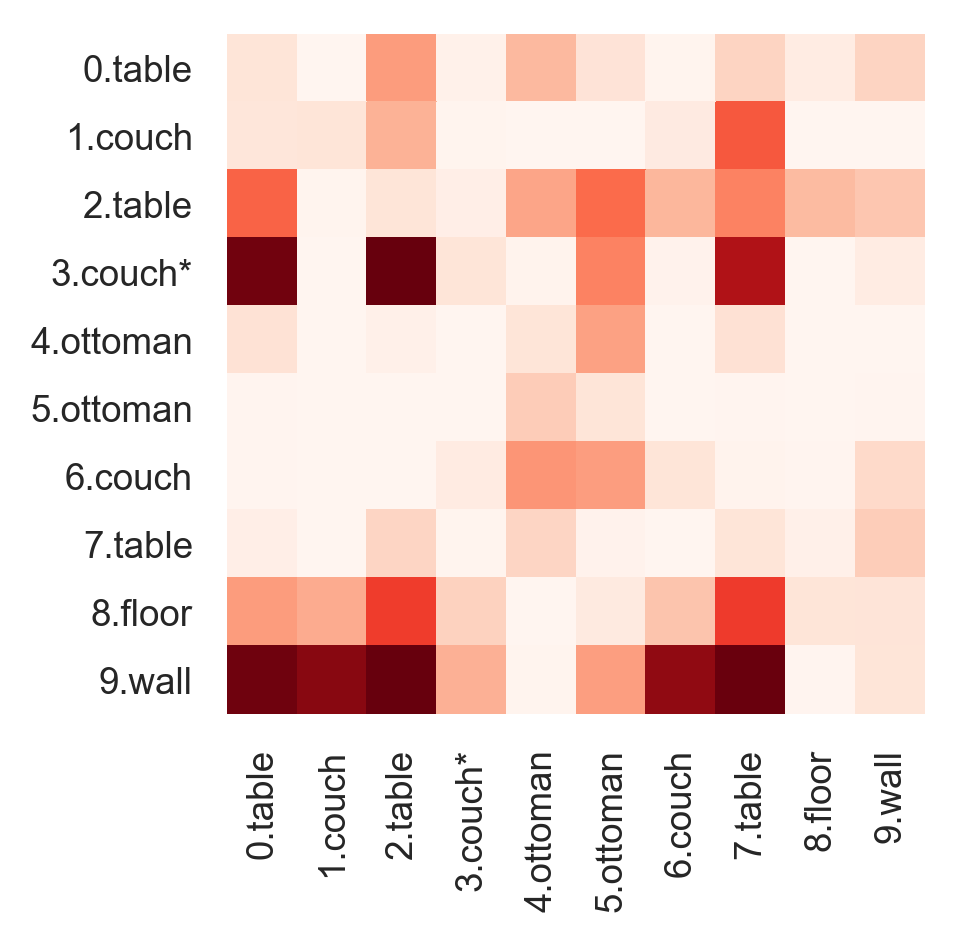}
    }
    \subfigure[RA-4 "middle"]{
        \includegraphics[width=4cm]{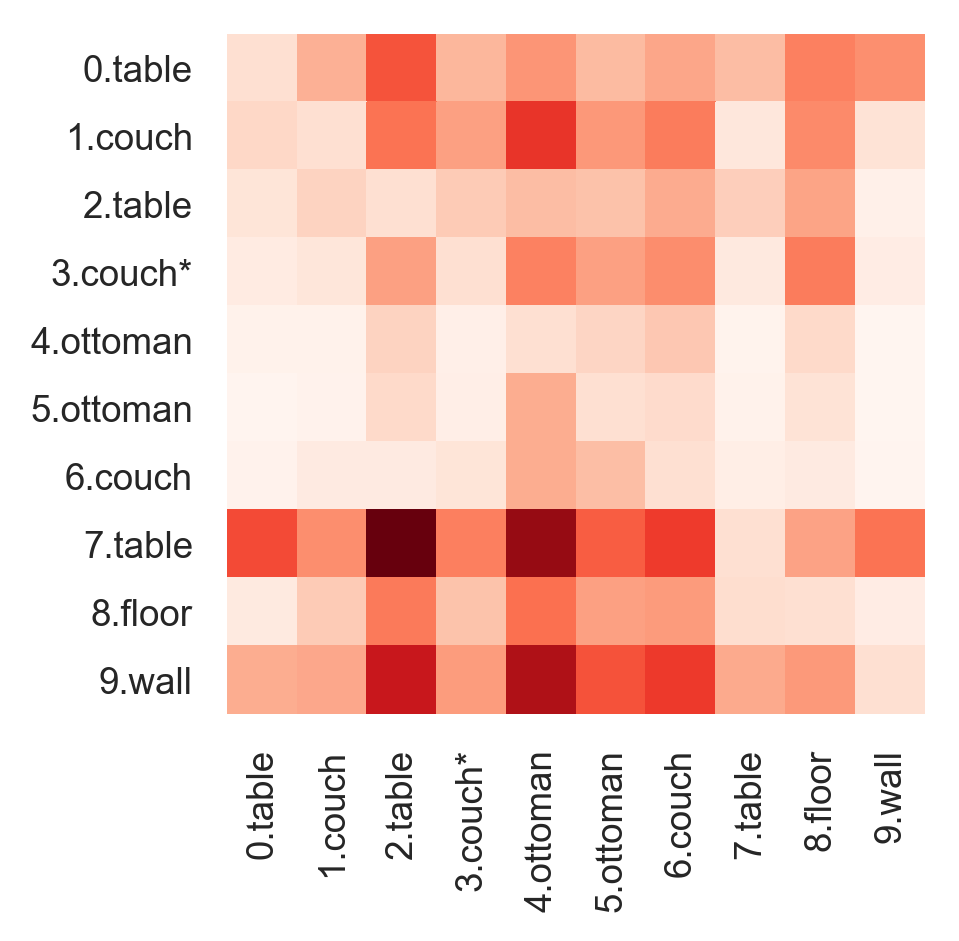}
         \label{fig:RA-4-middle}
    }
    \caption{Visualization of EA and RA. The attention maps are titled EA/RA-$x$ with $x$ denoting the layer depth. The third word "couch" marked with an asterisk denotes the referent. (a-d) Attention maps of EA. For each EA module, we report attention maps from 2 of the 4 heads. (e-h) Attention maps of RA responding to the words "between" and "middle".}
    \label{fig:attn_vis}
\end{figure*}

\subsection{Qualitative Analysis}

\subsubsection{Case Study}

We render the 3D scenes sampled from Nr3D test set and visualize the visual grounding results of our model. Figure~\ref{fig:case-study} shows part of the results.  Compared with ReferIt3DNet, our TransRefer3D performs better on both simple and hard cases because of a finely multi-modal context modeling. The first column in the figure indicates that, even in a complicated scene with a great number of distractors (5 monitors are replaced closely on a table), our model can still identify the referent from the context.

\subsubsection{Attention visualization of EA and RA}

Figure~\ref{fig:attn_vis} shows the visualization of our proposed attention modules enhancing visual features. In this case, the referring expression is "Middle part of couch, seams end in between the 2 beige cocktail tables". For both EA and RA, we visualize the attention maps extracted from the first and the last ERCB (represented as EA/RA-1/4 in the figure).

According to the EA-4 maps showed in Figure~\ref{fig:EA-4-head-1} and Figure~\ref{fig:EA-4-head-4}, it is very clear that in the last layer, EA matches the entity words with corresponding visual features (the couches and the tables in this case). We also observe that another EA head in the last layer matches the visual and linguistic features for couches and tables across. This behavior of EA perhaps helps establish a more expressive entity aware feature for the later grounding. 
See EA-1 maps in Figure~\ref{fig:EA-1-head-1} and Figure~\ref{fig:EA-1-head-2}, EA behaviour in the first layer also maps tables and couches to different modality but not such obviously, compared with the EA-4 maps.

In this case, we investigate RA responding to two relational words, "between" and "middle". The attention maps of RA show a similar phenomenon that the maps from the last layer are more understandable than ones from the first layer. See Figure~\ref{fig:RA-4-between}, the  \textit{0.table-4.ottoman} and \textit{0.table-5.ottoman} relation representations strongly respond to the word feature "between". This makes sense since the table is exactly located between the two ottomans in the scene. Notice that "between" is a ternary relation describing a spatial state of an entity relative to the other two entities. 
In Figure~\ref{fig:RA-4-middle}, the referred part of couch (id. 3) is located in the "middle" of most of strongly responded object pairs (e.g. \textit{7.table-2.table}, \textit{9.wall-2.table}, \textit{9.wall-4.ottoman}). The model learns to find objects surrounding the referent.
This instances imply that the proposed RA module can also capture relations more complicated than binary relations.

\section{Conclusion}

We propose TransRefer3D consisting of ERCBs which integrate EA and RA modules for entity-and-relation aware multimodal context modeling. The proposed model performs effectively on fine-grained 3D visual grounding. On both Nr3D and Sr3D, the model significantly achieves superior performance compared with previous approaches. This proves that Transformer is suitable to connect language and 3D vision and demonstrates the effectiveness of our entity-and-relation aware modeling. Hence, the proposed EA and RA modules and ERCB may have a positive influence to future works on 3D multimodal machine learning. In the future, we will further dig into Transformers for 3D modal tasks. As 3D vision is rapidly developing today, we believe that the proposed TransRefer3D will become an important reference.

\section{Acknowledgement}

This research is supported in part by National Natural Science Foundation of China (Grant 61876177), Beijing Natural Science Foundation (4202034), Fundamental Research Funds for the Central Universities.

\bibliographystyle{ACM-Reference-Format}
\balance 
\bibliography{sample-base}

\end{document}